\documentclass[sigconf, review=False]{acmart}
\usepackage{booktabs} % For formal tables
\usepackage{booktabs}
\usepackage[table]{xcolor}
\usepackage{colortbl}

\definecolor{lightgray}{gray}{0.9}
\definecolor{lightblue}{rgb}{0.88,0.93,0.95}
\definecolor{lightgreen}{rgb}{0.9,1,0.9}

\usepackage{tabularx}

\setcopyright{rightsretained}
\usepackage{subcaption}

\newcommand{\system}{NetraLink}

\acmDOI{10.1145/3774521.3774575}

\acmConference[ICVGIP'25]{16th Indian Conference on Computer Vision, Graphics and Image Processing}{December 2025}{Mandi, India}
\acmYear{2025}
\copyrightyear{2025}

\acmPrice{15.00}

\begin{document}
\title{Are We There Yet? Exploring the Capabilities of MLLMs \\in Assistive AI Applications}

\author[]{Shayon Dasgupta}
\affiliation{%
    \institution{IIT BHU}
    \country{India}
}

\author[]{Avijit Dasgupta}
\affiliation{%
    \institution{CVIT, IIIT Hyderabad}
    \country{India}
}
% \email{avijit.dasgupta@research.iiit.ac.in}

\authornote{Corresponding author}
% \email{avijit.dasgupta@research.iiit.ac.in}

\author[]{C. V. Jawahar}
\affiliation{%
    \institution{CVIT, IIIT Hyderabad}
    \country{India}
}

% \email{jawahar@iiit.ac.in}

% \title{Are We There Yet? Towards Assistive AI Applications with MLLMs}
\titlenote{Produces the permission block, and
  copyright information}

% \author{Submission Id 182}
% \affiliation{%
%   \institution{XYZ}
%   \streetaddress{XYZ}
%   \city{XYZ}
%   \state{XYZ}
%   \country{XYZ}
%   \postcode{000000}
% }

\renewcommand{\shortauthors}{}
\begin{abstract}

Multimodal Large Language Models (MLLMs) have redefined visual understanding by combining vision encoders with large-scale language models. This unified architecture enables strong performance on tasks like image captioning, visual question answering, and multimodal dialogue, often in zero- and few-shot settings. Their general-purpose capabilities and flexible interfaces make MLLMs a promising foundation for real-world vision-language applications.

Assistive AI aims to help users interact with their environments through natural language. These scenarios demand robust visual recognition, contextual reasoning, and multilingual comprehension—capabilities that MLLMs are believed to offer. However, their effectiveness in assistive settings remains to be fully understood.

In this work, we explore whether MLLMs can support Assistive AI by evaluating state-of-the-art models on real-world tasks: recognizing everyday objects like currency, answering questions based on scene text, and reading visually presented content across multiple languages. To this end, we developed a system, \system, using a head-mounted GoPro to capture real-world egocentric data, and collected a benchmark covering these assistive scenarios. Our findings provide a comprehensive diagnostic of current MLLMs, highlighting their strengths and limitations in enabling assistive technologies grounded in visual perception and language interaction. 
\end{abstract}



\maketitle

\section{Introduction}
\label{sec:intro}

Assistive AI systems aim to support individuals in navigating the world more independently, particularly those with visual, cognitive, or physical impairments. These technologies seek to provide real-time understanding of the user’s environment and communicate relevant information in an accessible manner. To be effective, Assistive AI must combine robust perception, high-level reasoning, and natural interaction—enabling capabilities such as object recognition, spatial understanding, and task-relevant feedback. Notably, India alone is home to over 4.95 million people with visual impairment and nearly 35 million individuals with disabilities and 0.24 million blind children overall~\cite{mannava2022current}, underscoring the widespread need for scalable and inclusive assistive solutions.

%While Assistive AI is often discussed in the context of visual impairment, its scope is far broader—extending to support for cognitive challenges, mobility assistance, and context-aware decision-making. 

Human visual attention is naturally drawn to high-level semantic elements, and prior eye-tracking studies have shown that text is one of the most consistently fixated regions in natural scenes—second only to faces~\cite{bylinskii2016should, wang2012attraction, cerf2009faces}. In particular, Cerf et al.~\cite{cerf2009faces} found that during free-viewing, people looked at text regions over 11× more often than control areas matched for size and location, underscoring the inherent salience of text in our visual environment. Importantly, participants struggled to avoid looking at text even when instructed to do so, suggesting that textual content is not only informative but also difficult to ignore. This has profound implications for Assistive AI: for individuals with visual or cognitive impairments, much of the crucial information encountered in daily life—signs, instructions, labels, or digital content—is locked behind text. Real-time text understanding can help users reconnect with their surroundings.

Motivated by the importance of visually presented text in human attention and daily functioning, we turn to recent advances in MLLMs~\cite{Vasu_2025_CVPR, liu2023llava, liu2024llavanext}, which have dramatically expanded the capabilities of modern vision systems by enabling them to interpret and reason over visual content through natural language. These models combine the powerful language understanding abilities of Large Language Models (LLMs) with high-capacity visual encoders, allowing them to process complex visual scenes and generate rich, contextually grounded responses. As a result, MLLMs have shown impressive performance across a wide spectrum of tasks, including image captioning~\cite{Peng_2025_CVPR}, object detection~\cite{yin2025rod}, segmentation~\cite{lai2024lisa}.%, and visual question answering~\cite{chen2024lion}.

Among the most promising application areas for MLLMs is Assistive AI, which demands real-time visual understanding, language-based reasoning, and seamless interaction to augment human capabilities. These systems can be transformative for individuals with disabilities, supporting tasks such as navigating unfamiliar environments, reading text, identifying objects, and following instructions. This shift in vision-language modeling opens exciting possibilities for building more adaptive and user-aligned AI. Yet, it remains unclear whether current MLLMs are ready for real-world assistive use, where robustness, relevance, and responsiveness are essential.

To explore the real-world applicability of MLLMs in Assistive AI, we develop a prototype system, \system, that simulates an egocentric assistive setup using a head-mounted GoPro\footnote{\url{https://gppro.in/product/gopro-hero10-black/}} camera. The system is capable of capturing images on user instruction and routing them through an MLLM / LLM remotely hosted on a laptop. This flexible, offline architecture enables real-time, interactive tasks without reliance on cloud infrastructure—making it suitable for practical, privacy-aware deployments. Using this setup, we systematically study the capabilities and limitations of existing MLLMs across several assistive scenarios common to our day-to-day lives.

% We categorize our evaluation into three core use cases. (a) Blind assistance: covering tasks like currency recognition, navigation, and scene text question-answering from complex environments; (b) Multilingual understanding: enabling menu card reading in Indic and English languages, crucial for accessibility in multilingual regions; (c) Reading comprehension: where the system observes a user reading a book for an extended period and is later queried to recall or summarize information, testing long-range grounding and memory. For each of these, we collect and annotate a custom dataset reflecting egocentric, real-world interactions. We benchmark multiple open-sourced MLLMs on this data to evaluate their alignment with assistive requirements, analyzing their strengths, failure modes, and readiness for deployment in everyday assistive scenarios.

We categorize our evaluation into three core use cases: (a) Blind assistance – tasks like currency recognition, navigation, and scene-text question answering in complex environments; (b) Multilingual understanding – reading menu cards in Indic and English, crucial for accessibility in multilingual regions; (c) Reading comprehension – where the system observes a user reading a book over an extended period and is later queried to recall or summarize information, testing long-range grounding and memory. For each, we curate and annotate a custom dataset capturing egocentric, real-world interactions. We benchmark multiple open-source MLLMs to assess their alignment with assistive needs, analyzing strengths, failure modes, and practical readiness for everyday deployment.

In summary, this work presents the first systematic study of MLLMs for egocentric Assistive AI applications. Our key contributions are as follows:

\begin{itemize}
\item We develop a head-mounted assistive AI system, \system, to simulate real-world data capture and interaction in egocentric assistive scenarios for evaluating the capabilities of MLLMs.
\item We define five practical assistive tasks—scene-text recognition, currency-note recognition, navigation assistance, multilingual menu reading and interpretation, and long-form content reading.
\item We collect and annotate a novel egocentric dataset spanning diverse real-world assistive scenarios.
\item We benchmark multiple MLLMs on our dataset to evaluate their effectiveness in assistive contexts and highlight their failure modes.
\end{itemize}

\section{Related Works}

\paragraph{Egocentric Vision.}
Egocentric vision—where visual data is captured from a first-person viewpoint—is gaining traction in the computer vision community~\cite{damen2018scaling, grauman2022ego4d, grauman2024ego}. This perspective offers unique advantages for understanding user interactions, capturing contextual cues like gaze, motion, and hand-object contact. These characteristics make egocentric data particularly valuable for Assistive AI applications, as they closely mirror how users perceive and experience their surroundings. While large-scale benchmarks such as Ego4D~\cite{grauman2022ego4d} and Ego-Exo4D~\cite{grauman2024ego} have catalyzed progress on tasks like action recognition and object tracking, they remain limited in their coverage of practical assistive scenarios—especially those involving fine-grained visual reasoning, multilingual text understanding, and real-time human-in-the-loop interaction. To address this gap, we develop a new egocentric dataset explicitly focused on everyday assistive use cases grounded in real-world visual challenges.

% \paragraph{Multi-Modal Large Language Models.}
% MLLMs~\cite{zhang2023gpt, Vasu_2025_CVPR, liu2023llava, liu2024llavanext} combine vision encoders with large language models to support multimodal reasoning and generation. By aligning visual and textual modalities in a unified framework, they offer strong generalization across diverse tasks, from image captioning to visual question answering~\cite{Peng_2025_CVPR, yin2025rod, lai2024lisa, chen2024lion}. Their instruction-driven nature removes the need for task-specific pipelines, making them a promising foundation for assistive systems. Yet, their robustness and utility in egocentric, real-world settings remains underexplored.

\paragraph{Multi-modal Large Language Models}
MLLMs~\cite{zhang2023gpt, Vasu_2025_CVPR, liu2023llava, liu2024llavanext} extend LLMs with the ability to perceive, reason, and generate across multiple modalities. They typically comprise a modality encoder, a pre-trained LLM, and a projector that aligns representations across modalities. A key milestone in this direction was CLIP~\cite{radford2021learning}, which introduced vision–language pretraining to learn a shared representation space for images and text, laying the groundwork for unified multimodal reasoning. Building on this foundation, recent MLLMs have achieved strong results in diverse tasks such as image captioning~\cite{Peng_2025_CVPR}, object detection~\cite{yin2025rod}, segmentation~\cite{lai2024lisa}, and visual question answering~\cite{chen2024lion}. By integrating perception and language understanding within a single framework, MLLMs eliminate the need for highly specialized task-specific architectures, offering a scalable, instruction-driven approach to vision problems.

Despite these advances, it remains uncertain how close we are to realizing general-purpose personal AI assistants that can operate robustly in day-to-day life. While many individual tasks have seen substantial gains, comprehensive evaluations of MLLMs in realistic assistive scenarios—particularly those involving egocentric visual data—are still lacking. In this work, we address this gap by systematically assessing state-of-the-art MLLMs on a set of everyday assistive use cases.

\paragraph{Text Recognition and Understanding.}
Text is a rich source of semantic information in our visual world, appearing on signs, screens, product labels, and documents. Accurate recognition and interpretation are crucial for tasks like navigation, reading instructions, and object identification—especially for individuals with visual impairments. Early OCR systems such as Tesseract~\cite{smith2007overview} and EAST~\cite{zhou2017east} have evolved into deep learning–based models like CRNN~\cite{shi2016end}, ASTER~\cite{ShiWLYB16}, CRAFT~\cite{baek2019character}, and DBNet~\cite{liao2020real}, while document understanding approaches such as LayoutLMv3~\cite{huang2022layoutlmv3}, Donut~\cite{kim2022donut}, and DocFormer~\cite{appalaraju2021docformer} enable reasoning over structured layouts. More recently, multi-modal large language models (MLLMs) like BLIP-2~\cite{li2023blip}, Kosmos-2~\cite{peng2023kosmos}, and GPT-4V~\footnote{\url{https://openai.com/contributions/gpt-4v/}} have shown strong zero-shot text-grounded reasoning.

Benchmarks such as DocVQA~\cite{mathew2021docvqa}, SceneText-VQA~\cite{biten2019scene}, COCO-Text~\cite{veit2016coco}, and TextVQA~\cite{singh2019towards} have advanced the field, but mostly focus on static or curated images. These settings fail to capture the challenges of egocentric vision—where text may be occluded, motion-blurred, or context-dependent. Moreover, traditional OCR with rule-based NLP pipelines may yield higher accuracy and lower latency for simple text extraction, they are brittle and fail on open-ended, context-dependent queries. In contrast, MLLMs provide a unified, end-to-end approach that couples perception with language reasoning—enabling flexible, conversational, and intelligent interaction. Our work fills this gap by evaluating recognition and understanding in real-world egocentric assistive scenarios using both OCR- and MLLM-based methods. While Karamolegkou \textit{et al.}~\cite{karamolegkou2025evaluating} take a first step, a thorough assessment of MLLMs in multilingual, noisy, and navigation-oriented settings has been missing—precisely what we provide in this paper.

% Text is a pervasive and semantically rich element in our visual environment, appearing on signs, screens, labels, and documents. Recognizing and interpreting this text is critical for everyday tasks such as navigation, reading instructions, or identifying objects—especially for individuals with visual impairments. Benchmarks such as DocVQA~\cite{mathew2021docvqa}, SceneText-VQA~\cite{biten2019scene}, COCO-Text~\cite{veit2016coco}, and TextVQA~\cite{singh2019towards} have pushed the field toward more complex reasoning over embedded text. However, most existing datasets focus on static or curated views, falling short of modeling the challenges in egocentric scenarios—where text may be occluded, distorted, or context-dependent. Our work addresses this gap by evaluating text understanding in real-world, egocentric assistive settings.
\section{\system: System Overview}
\label{sec:architecture}

\begin{figure}[!t]
    \centering
    \includegraphics[width=\linewidth]{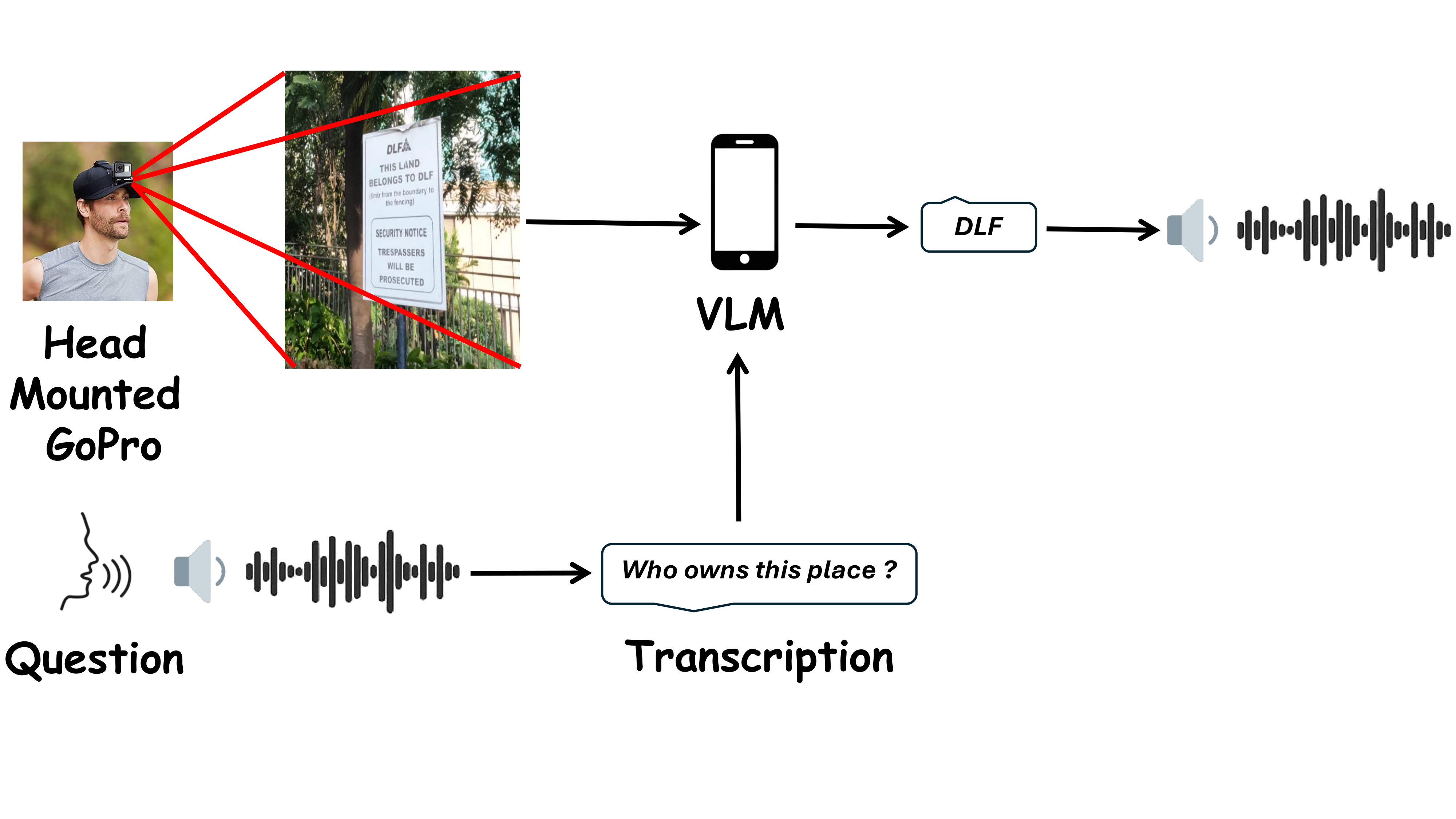}
    \caption{Overview of our Assistive AI system, \system.}
    \label{fig:system_architecture}
\end{figure}

Fig.~\ref{fig:system_architecture} illustrates the end-to-end components of our egocentric assistive AI prototype. The system is organized into five tightly integrated yet modular components: (1) Audio Front-end, (2) Speech-to-Text, (3) Image Acquisition, (4) Vision–Language Inference, and (5) Response Generation and Logging. This modularity not only clarifies the dataflow but also enables independent development, testing, and future substitution of individual modules to accommodate advances in each subfield.

% \begin{figure}[ht]
%   \centering
%   %\includegraphics[width=\linewidth]{figures/system_architecture.pdf}
%   \caption{Overview of the proposed system architecture.}
%   \label{fig:system_architecture}
% \end{figure}

\paragraph{Audio Front-end} This module continuously monitors ambient sound and detects when the user explicitly invokes the system. We use a wake-word detector on short, overlapping audio snippets (e.g., 3 s), combining lightweight frame-based energy analysis with a voice-activity detector (VAD)\footnote{\url{https://github.com/wiseman/py-webrtcvad}}. This two-stage design reduces false triggers from background noise or unrelated speech, conserving computation and protecting privacy. When the wake-word is recognized, the module signals readiness and switches to active query capture, again using the same VAD in streaming mode: it buffers audio frames until it observes sustained silence beyond a configurable threshold (e.g., 1 s of non-speech). This silence-based segmentation adapts to speaking rate and ambient conditions, capturing the full utterance without manual start/stop. A small circular buffer is maintained internally to avoid clipping the start of the query, improving robustness in real-world use.

\paragraph{Speech-to-Text} Once the audio front-end produces a segmented utterance, it is sent to the Speech-To-Text (STT) module for transcription using a locally hosted or on-device recognizer invoked via a simple command-line interface. The STT engine applies a state-of-the-art acoustic and language model stack to perform noise reduction, phoneme decoding, and punctuation insertion, converting waveforms into text. The output is then normalized—tokenized, lower-cased, and trimmed—before being passed to the next stage. Because STT runs as a separate subprocess, we can swap or update transcription backends without affecting capture or reasoning modules. In practice, our STT module is robust, achieving an average word error rate (WER) of approximately 7\% in noisy environments across 50 samples.

\paragraph{Image Acquisition} After transcription, the system immediately captures the visual context relevant to the user’s request. A head-mounted, user-facing camera provides a continuous live view but records only a single high-resolution image when triggered. We interface with it through a lightweight API that exposes a simple `capture()` call. Each frame is time-stamped and given a unique identifier based on the capture epoch, then stored in a structured directory hierarchy organized by date and session. This keeps the camera unobtrusive—remaining in standby until explicitly activated—while conserving battery life and limiting unnecessary data capture. Storing each image alongside its corresponding transcript also provides precise cross-modal alignment for error analysis, user studies, and future module retraining.

\paragraph{Vision-Language Inference} At the core of the system is the vision–language inference engine. We build a single multimodal prompt by combining the user’s transcribed query with a reference to the captured image, following a template such as: \begin{quote} \footnotesize \texttt{“Image Context: [image\_ID] User Query: [transcribed\_text]”} \end{quote}
This prompt conditions both the visual encoder and the language model. The engine runs as a standalone command-line process, with arguments specifying the model checkpoint, image path, prompt string, and generation parameters (e.g., maximum tokens, temperature). We measure latency from process launch to response completion for profiling and optimization. Decoupling inference into its own executable cleanly separates concerns and lets us swap architectures, upgrade to newer foundation models, or tune hyperparameters without modifying the audio front-end.% or response pipeline.

\paragraph{Response Generation and Logging}
The raw text from the vision–language engine is first cleaned to remove system artifacts (e.g., special tokens) and then logged to persistent storage. For our prototype, we use a simple spreadsheet (Excel), appending rows with image ID, user query, model response, inference latency, and timestamp. This structured log supports both quantitative benchmarking and qualitative inspection. For real-time feedback, the cleaned response is sent to the text-to-speech (TTS) module, which synthesizes a natural-sounding waveform. The TTS engine is also invoked as an external process, with configurable voice and language settings, and the resulting audio is played through the headset speakers. We store both the intermediate text and the generated audio in a dedicated directory, enabling later user studies and ensuring full reproducibility.

\paragraph{Design Highlights and Extensibility}
The system is built from modular components with clean interfaces, making it easy to swap wake-word detectors, STT engines, or inference backends. All processing runs locally to preserve privacy, with no cloud communication. Wake-word gating and silence-based segmentation keep the system efficient and responsive, even on limited hardware. A single configuration file controls key parameters (e.g., VAD thresholds, wake-word sensitivity, inference settings), simplifying tuning and deployment. Note that system-level optimizations such as model compression, knowledge distillation, and efficient architectures are beyond the scope of this paper, as our primary objective is to first establish and analyze the fundamental limitations of MLLMs in assistive scenarios. We consider these optimisations as natural next steps toward deployable systems.

\section{Benchmark Setup}
\label{sec:app}
% In this section, we outline X practical applications of the developed head-mounted assistive AI system and describe the associated datasets.
\begin{table}[t]
\centering
\setlength{\tabcolsep}{2pt}
\renewcommand{\arraystretch}{1}
\begin{tabularx}{\linewidth}{l X X}
\toprule
\textbf{Task} & \textbf{Sample Size} & \textbf{Source of data} \\
\midrule
1 & 8025 images & 133 min of egocentric video \\
2 & 150 images & Banknotes from 5 countries\\
3 & 3 environments & 133 min of egocentric videos \\
4 & 72 menu images & In-the-wild English, Bengali, Hindi menu photos \\
5 & 37 page-level images& 62 min of egocentric storybook reading  \\
\bottomrule
\end{tabularx}
\caption{Summary of the five tasks in our dataset. Extending the dataset with additional tasks is part of future work.}
\label{tab:dataset_tasks}
\end{table}

In this section, we outline \textit{five} practical tasks of the developed head-mounted assistive AI system—each centered around real-world text understanding tasks as motivated in Section~\ref{sec:intro}—and describe the associated datasets used to evaluate these capabilities. We provide an overview of all tasks and dataset statistics in Table~\ref{tab:dataset_tasks}.

\begin{figure}[!t]
    \centering
    \includegraphics[width=\linewidth]{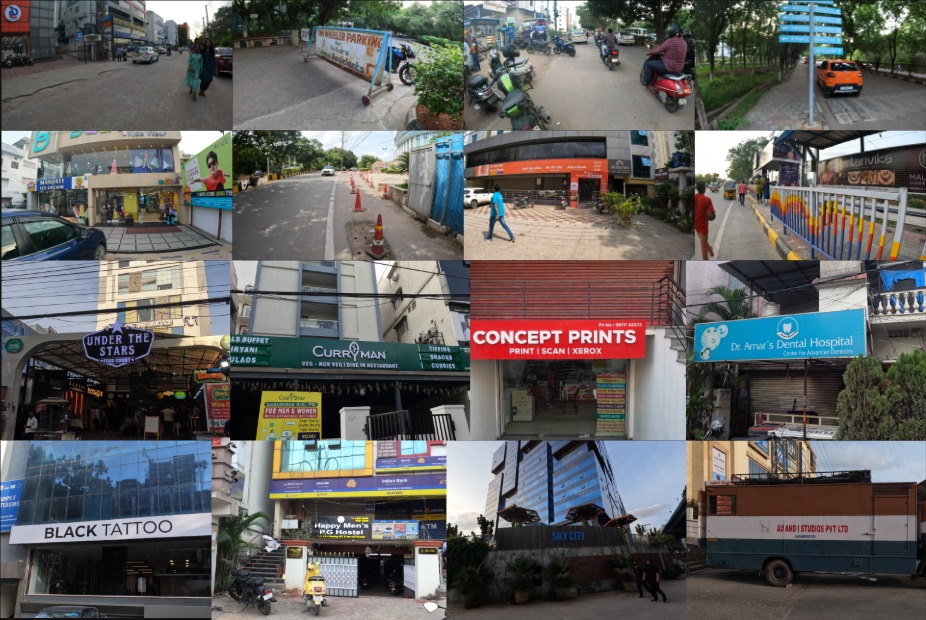}
    \caption{Examples of egocentric images used to evaluate MLLMs for scene-text recognition, captured using our head-mounted assistive AI system across diverse urban environments—including sidewalks, pedestrian crossings, market areas, and residential streets. These images feature varied instances of scene text such as signs, storefronts, advertisements, and labels, which are vital for real-world understanding and navigation.}
    \label{fig:scene-text}
\end{figure}

% \subsection{Blind Assistance}
% \label{subsec:blind-assistance}
Globally, over 2.2 billion people experience near or distance vision impairment~\cite{fricke2018global, burton2021lancet}, including an estimated 43.3 million who are blind~\cite{vision2024global}. This widespread prevalence highlights the critical need for effective assistive technologies that can support blind and visually impaired individuals in navigating daily challenges. To help address these needs, we focus on three key tasks as follows - 

%Vision loss can severely hinder access to information, financial independence, and educational resources—areas where inclusive design and intelligent systems can make a transformative difference.

\noindent\textbf{Task 1: Scene-Text Recognition.}
Text is one of the most important visual elements in our daily lives~\cite{bylinskii2016should, wang2012attraction, cerf2009faces}. 
Studies show that people consistently focus on text in natural scenes—often even more than on other visual cues—due to its high-level semantic value. Whether it's a street sign, product label, or digital interface, written text helps individuals interpret and interact with their environment. For blind individuals, this information remains largely inaccessible, making navigation, object recognition, and understanding instructions challenging. Real-time scene-text recognition can bridge this gap, enhancing independence and situational awareness.

We frame this as a visual question answering (VQA) task, where both questions and answers are in natural language. Using our \system, we captured 133.31 minutes of egocentric video across varied urban settings with a head-mounted GoPro, sampling frames at 1 FPS to obtain 8,025 images. Example questions include “What does the text on the sign say?” or “Which shop is this?”—assessing a model’s ability to detect, read, and reason about text in context (see Fig.~\ref{fig:scene-text}).

\begin{figure}[!t]
    \centering
    \includegraphics[width=\linewidth]{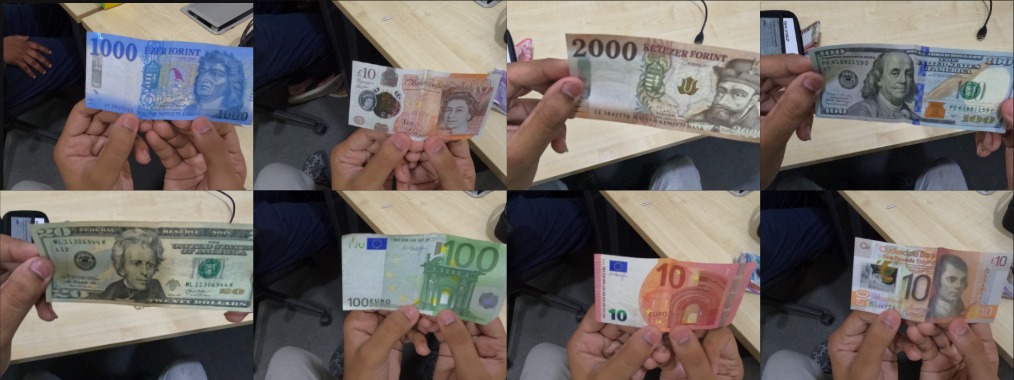}
    \caption{Examples of currency-note images used to evaluate MLLMs across five countries: India, the USA, the UK, Europe, and Hungary. Each note is paired with multiple question types—such as “What is the value of this note?”—to assess the model's ability to recognize denominations and reason about visual financial cues under diverse layouts, languages, and designs.}
    \label{fig:currency}
\end{figure}

\noindent\textbf{Task 2: Currency-Note Recognition.}
The goal of this task is to assess a model’s ability to answer questions about paper currency across diverse designs. For blind individuals, banknote identification is challenging due to the lack of consistent tactile markers, making transactions—such as paying, receiving change, or verifying denominations—difficult and error-prone. Real-time currency recognition can enhance financial independence and confidence in economic interactions.

We frame this as a VQA task, with both questions and answers in natural language. Our dataset contains 150 images of banknotes from five countries—India, the USA, the UK, Europe, and Hungary—captured to represent varied designs and denominations. For each of the 30 images per country, we created four question types (e.g., “What is the origin of this note?”) to evaluate recognition accuracy and reasoning over currency-specific visual features (see Fig.~\ref{fig:currency}). This design enables cross-cultural, denomination-aware evaluation and encourages robustness in real-world financial scenarios.

\begin{figure}[!h]
    \centering
    \includegraphics[width=\linewidth]{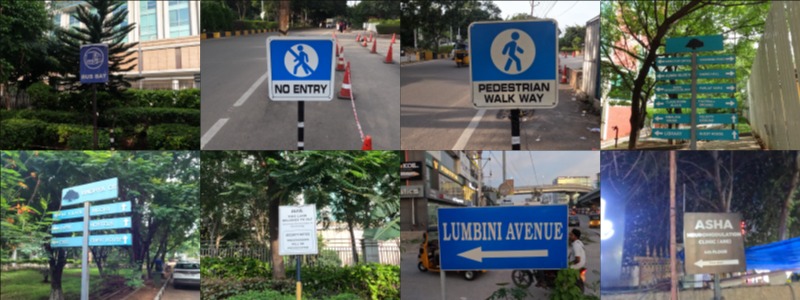}
    \caption{Sample egocentric images captured in varied outdoor environments—including a university campus, residential neighborhood, and commercial market—used to evaluate the navigation capabilities of MLLMs in dynamic, real-world settings.}
    \label{fig:navigation}
\end{figure}

\noindent\textbf{Task 3: Navigation Assistance.} This task evaluates a model’s ability to interpret egocentric scenes for real-time, language-guided navigation. Moving through unfamiliar or dynamic environments is particularly challenging for visually impaired individuals, as critical cues—such as signs, crossings, and pathways—are primarily visual. While canes and GPS-based audio tools aid mobility, they provide limited spatial and semantic context. Vision-language systems can bridge this gap by offering context-aware guidance, improving both safety and independence.

The evaluation is posed as a natural language question–answering problem grounded in visual input. Using our \system, we recorded egocentric video in three outdoor settings: a university campus, a residential neighborhood, and a commercial market. From this footage, we extracted high-resolution frames covering diverse mobility scenarios. A subset was annotated with navigation-focused questions such as “What is in front of me?”, “Is there a crossing ahead?”, and “Can I turn right here?” to assess spatial reasoning and scene interpretation (see Fig.~\ref{fig:navigation}).

% \subsection{Understanding Indic Languages}
% Understanding visual information in multiple languages is a challenge not only for people with visual impairments but also for the general population. Public-facing materials such as menus, signboards, and official documents are often written in local languages or in a mix of scripts. Visitors, tourists, and even locals may struggle when confronted with unfamiliar scripts or multilingual layouts. This makes robust multilingual visual understanding a critical capability for both assistive and general-purpose AI systems. To evaluate the capabilities of current MLLMs in this regard, we design the following task.
Understanding visual information across multiple languages is difficult for both people with visual impairments and the general population. Public materials like menus, signboards, and official documents often mix scripts and languages, causing confusion even for locals. This makes robust multilingual visual understanding a key requirement for assistive and general-purpose AI, motivating the following task.

\noindent\textbf{Task 4: Multilingual Menu Card Reading and Interpretation.}
Restaurant menus are a prototypical example of complex multilingual content, with diverse layouts, decorative fonts, and mixed-language text (e.g., dish names in Hindi or Bengali with English descriptions). For someone who cannot clearly see the menu or does not know one of the scripts, this becomes a major barrier. An assistive AI system should visually parse the menu, understand its content across languages, and answer natural-language questions.

\begin{figure}[!t]
    \centering
    \includegraphics[width=\linewidth]{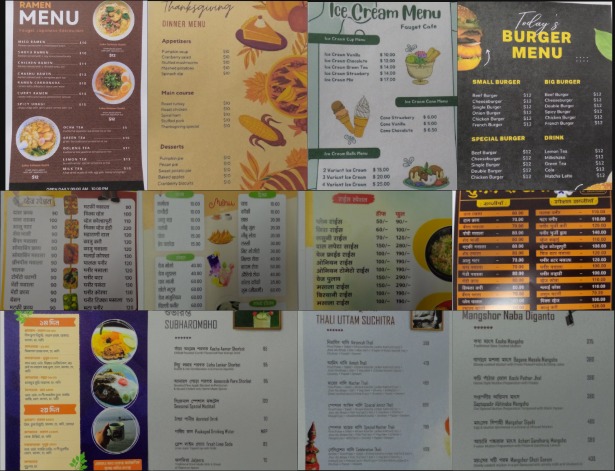}
    \caption{Sample images of restaurant menu cards in Hindi, Bengali, and English used to evaluate multilingual text understanding. The dataset includes printed menus in varied layouts and scripts, reflecting real-world diversity in visual and linguistic presentation.}
    \label{fig:menu}
\end{figure}

Using our \system~ platform, we build a head-mounted multimodal menu-card VQA setup. The task is to answer natural-language questions about menu content by jointly leveraging visual and textual cues across languages. We curate 72 real-world menu images (Fig.~\ref{fig:menu}): 18 in English, 28 in Bengali, and 26 in Hindi, covering single-language, bilingual, and mixed-script layouts. Captured under varied conditions, they vary in fonts, image quality, and background clutter. Each image is annotated with aligned question–answer pairs, enabling systematic evaluation of a model’s ability to extract, interpret, and reason over multilingual text in realistic scenarios.

% 18 English
% 28 Bengali
% 26 Hindi

% To assess the reasoning abilities of MLLMs, we crafted three distinct question templates, each probing a different aspect of menu comprehension:
% \begin{itemize}
% \item \emph{Price inquiries}: asking the cost of individual items.
% \item \emph{Category listings}: requesting any two items from a specified category (e.g., appetizers, beverages).
% \item \emph{Combined prompts}: querying for dishes under a thematic heading.
% \end{itemize}

% Long-form content understanding is essential for tasks that require integrating information across multiple observations, such as following narratives, tracking entities, and maintaining context over time. For assistive AI, this capability enables richer support in education, storytelling, and real-world reading assistance beyond surface-level text recognition.

% \subsection{Reading Long-form Content}

\begin{figure}[!t]
    \centering
    \includegraphics[width=\linewidth]{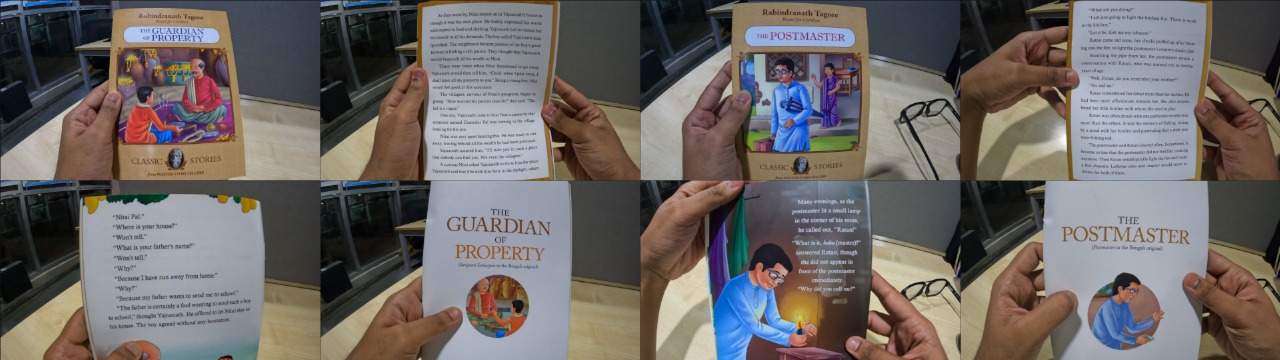}
    \caption{Sample page-level images from two printed storybooks captured using our head-mounted assistive AI system. These images are used to evaluate long-form narrative understanding through contextual question answering on characters, relationships, and plot elements.}
    \label{fig:bookqa}
\end{figure}

% Printed storybooks contain rich narrative structures that require more than surface-level text recognition—they demand contextual AI capable of tracking characters, events, and relationships across multiple pages. Unlike scene text or signage, book content involves deeper semantic understanding, making it a valuable benchmark for evaluating models' long-range reasoning and comprehension abilities. Building systems that can interpret such content expands access to literature and education, enabling use cases such as story-based question answering, summarization, and reading assistance in natural settings.

% \noindent\textbf{Task 5: Narrative Reading Comprehension.} To evaluate MLLMs on narrative understanding, we captured 62 minutes of egocentric video while reading \textit{two} printed storybooks using our head-mounted assistive AI system. From these recordings, we extracted 37 page-level images under natural reading conditions, reflecting challenges such as varied lighting, page curvature, and hand movement. To probe deeper comprehension, we designed a suite of narrative-focused question-answer pairs covering character identities, relationships, and plot developments. Example questions include: “Who is the miserly old man in the first story?”, “Why does Vrindavan Kundu quarrel with his father?”, and “What medical recommendation caused the dispute?” (see Figure~\ref{fig:bookqa}). This dataset enables evaluation of a model’s ability to perform long-form, context-aware reasoning from visual text inputs.

Understanding long-form content is a challenge that extends beyond assistive contexts, affecting anyone who needs to follow a narrative, track entities, or integrate information over time. Books, reports, and other extended texts require readers to maintain continuity across multiple observations rather than extracting meaning from isolated snippets. For AI systems, mastering this skill is critical for applications in education, storytelling, summarization, and real-world reading assistance—enabling richer, context-aware interaction with extended materials.

\noindent\textbf{Task 5: Reading Long-form Content.} Storybooks are a natural example of long-form visual text that demands continuity tracking and deeper reasoning. Unlike signage or short passages, where meaning can be grasped from a single frame, narratives in printed books require models to recognize and integrate information across multiple pages. This involves accurate OCR, entity and relationship tracking, and maintaining plot context—skills essential for delivering meaningful, context-aware answers to reader queries.

% Add in your preamble if not already present:
% \usepackage{booktabs,tabularx}

To evaluate this capability, we curated a dataset by recording 62 minutes of egocentric video while reading two printed storybooks using our head-mounted assistive AI system, \system. From these recordings, we extracted 37 page-level images under realistic reading conditions, which include challenges such as varied lighting, page curvature, partial occlusions from hands, and non-uniform alignment. Each page is annotated with narrative-focused question–answer pairs in natural language, requiring reasoning over one or more pages. The questions span character identification, relationship mapping, and plot development (e.g., “Who is the miserly old man in the first story?”, “Why does Vrindavan Kundu quarrel with his father?”, “What medical recommendation caused the dispute?”). This benchmark measures a model’s ability to couple OCR with long-range, context-aware reasoning, advancing both assistive story-based QA and general-purpose narrative comprehension. Some samples are shown in Fig.~\ref{fig:bookqa}.

\section{Evaluation and Analysis}

% \begin{figure*}[h]
%     \centering
%     \includegraphics[width=\textwidth]{images/scene-text-results.png}
%     \caption{Qualitative results for scene-text recognition using FastVLM~\cite{}. Correct predictions are highlighted in {\color{green}\textbf{green}}, while incorrect ones are marked in {\color{red}\textbf{red}}. FastVLM~\cite{} performs well when text is clear and centrally placed, but struggles in cases where the text is {\em blurred}, {\em very small}, or appears against {\em cluttered backgrounds}, likely due to limitations in spatial resolution and fine-grained attention.}
%     \label{fig:scene-text-labels}
% \end{figure*}

We evaluate a set of state-of-the-art MLLMs—including Gemma-3-4B~\cite{team2025gemma}, Gemma-3-12B~\cite{team2025gemma}, Gemma-3-27B~\cite{team2025gemma}, LLaVA~\cite{liu2023visual}, LLaVA-LLaMA3~\cite{hanoona2024LLaVA++}, LLaVA-Phi3-3.8B~\cite{hanoona2024LLaVA++}, Qwen2.5-VL-7B~\cite{bai2025qwen2}, and BakLLaVA~\cite{liu2024improved}—to analyze their effectiveness across five assistive AI applications (\textbf{Task} \textbf{1-5}) proposed in this work. Each task is cast as a question answering (QA) problem to ensure a consistent and unified evaluation framework across heterogeneous input types. This design enables a joint assessment of both visual recognition and language reasoning abilities. In line with previous works~\cite{mathew2021docvqa, biten2019scene}, we employ the Average Normalized Levenshtein Similarity (ANLS) score as our primary evaluation metric to quantify response accuracy. We also evaluate the LLM-as-a-judge~\cite{li2024llms} to evaluate the correctness of the answers. In addition to the ANLS score, we also recorded the average inference latency for each model to assess their practicality in real-time assistive applications, highlighting trade-offs between accuracy and responsiveness.

% \begin{figure*}[h]
%     \centering

%     \begin{subfigure}{\textwidth}
%         \centering
%         \includegraphics[width=0.95\textwidth]{images/scene-text-results.png}
%         \caption{Scene-text Recognition.}
%         \label{fig:scene-text-labels}
%     \end{subfigure}

%     \vspace{1em}

%     \begin{subfigure}{\textwidth}
%         \centering
%         \includegraphics[width=0.95\textwidth]{images/currency_results.png}
%         \caption{Currency Recognition}
%         \label{fig:currency_results}
%     \end{subfigure}

%     \caption{Qualitative examples showing strengths and limitations of different MLLMs across scene-text and currency recognition tasks. Correct predictions are in {\color{green}\textbf{green}}, errors in {\color{red}\textbf{red}}. Performance drops on blurred, small, or cluttered text.}
% \end{figure*}

% \subsection{Evaluation of Blind Assistance}
% \label{subsec:blind}

\begin{table}[h]
\small
\centering
\caption{\textbf{Quantitative results for Task 1:} Average latency and ANLS for each model across all scene-text queries, with the best and second-best performers highlighted in bold and \underline{underline}, respectively.}
\label{tab:scene_text_metrics}
\begin{tabular}{lrrr}
\toprule
\textbf{Model}         & \textbf{Latency (s)}  & \textbf{ANLS} & \textbf{LLM}\\
\midrule
LLaVA~\cite{liu2023visual}                   & 3.01            & 0.32         & 0.51   \\
Gemma-3-27B~\cite{team2025gemma}             & 19.32           & 0.35         & 0.78   \\
LLaVA–Phi3-3.8B~\cite{hanoona2024LLaVA++}    & \textbf{1.95}   & 0.36         & 0.45   \\
Gemma-3-12B~\cite{team2025gemma}             & 15.73           & 0.42         & 0.77   \\
Gemma-3-4B~\cite{team2025gemma}              & 13.37           & 0.44         & 0.71   \\
BakLLaVA~\cite{liu2024improved}              & \underline{2.17} & 0.47        & 0.43   \\
LLaVA–LLaMA3~\cite{hanoona2024LLaVA++}       & 2.65            & 0.50         & 0.50   \\
Qwen2.5-VL-7B~\cite{bai2025qwen2}            & 26.13    & \underline{0.57}    & 0.76   \\
FastVLM-7B~\cite{vasu2025fastvlm}            & 14.88       & \textbf{0.66}    & 0.85  \\

\bottomrule
\end{tabular}
\end{table}

%To assess the role of text in everyday interactions, we analyzed a 133.31-minute egocentric video (see Section~\ref{sec:app}) captured using our system. 

% Each frame was processed individually using the \texttt{IndicPhotoOCR}\footnote{\url{https://github.com/Bhashini-IITJ/IndicPhotoOCR}} engine on a mid-range ARM-based CPU in a single-threaded configuration. For each frame, we recorded inference time, the number of words recognized, and variations in processing duration—particularly under visually complex conditions such as dense signage or textured backgrounds—to identify potential worst-case latency spikes. A subset of frames was manually reviewed to validate text-region coverage and recognition accuracy, ensuring pipeline consistency.

% Across the 8,025 frames, the OCR system produced 119,791 recognized words, corresponding to a mean of 14.93 words per frame ($\sigma \approx 3.4$). The average inference time was 14.29 seconds per frame ($\sigma \approx 2.1$), with peaks reaching 18.7 seconds in cluttered scenes and a minimum of 9.8 seconds in sparse-text frames. These measurements establish both an average and variability baseline relevant for assessing real-time feasibility. Despite mixed lighting and motion conditions, memory usage remained stable, indicating the suitability of the selected OCR engine and hardware configuration for deployment on embedded platforms. However, the observed latency variability highlights the need for adaptive scheduling or prioritized frame selection in time-sensitive assistive applications.

\begin{figure}[!ht]
    \centering
    \includegraphics[width=\columnwidth]{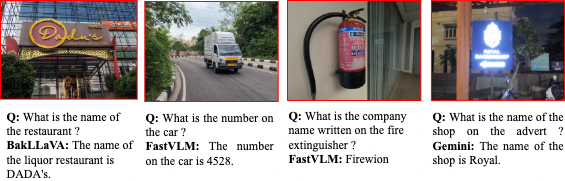}
    \caption{Qualitative examples of failure cases for scene-text recognition.}
    \label{fig:str-errors}
\end{figure}

\noindent\textbf{Task 1: Scene-Text Recognition.} 
Text is integral to daily life — appearing on signs, storefronts, menus, and instructions—making its recognition vital for assistive technologies. To show this, we processed 8,025 video frames (113 minutes of footage) using the \texttt{IndicPhotoOCR}\footnote{\url{https://github.com/Bhashini-IITJ/IndicPhotoOCR}} engine on a mid-range ARM CPU, recording word counts and variability under cluttered and sparse-text scenes. The system recognized 119,791 words (14.93/frame), demonstrating the significant volume of textual information available in everyday environments and the importance of accurate extraction for real-world assistive applications.

To assess the suitability of modern MLLMs for assistive reading in real-world settings, we evaluated nine models on a benchmark of natural images with printed and handwritten text under cluttered, low-contrast, and occluded conditions. FastVLM-7B~\cite{vasu2025fastvlm} achieved the highest ANLS (0.66), followed by Qwen2.5-VL-7B~\cite{bai2025qwen2} (0.57), while Gemma-3-4B~\cite{team2025gemma} (0.44) outperformed larger variants, showing that parameter scale alone does not ensure fine-grained text understanding. For latency-sensitive scenarios, LLaVA–Phi3-3.8B~\cite{hanoona2024LLaVA++} (0.36 ANLS, 1.95s) and BakLLaVA~\cite{liu2024improved} (0.47 ANLS, 2.17s) provided the best accuracy–speed balance, underscoring the importance of vision–language alignment and task-specific grounding over model size. This is consistent with typical scaling trends, inference latency increases with LLM parameter count. FastVLM-7B~\cite{vasu2025fastvlm} likewise attains the top LLM-as-a-judge~\cite{li2024llms} score, underscoring its strong text understanding and reasoning capabilities.

Our analysis of failure cases (see Fig.~\ref{fig:str-errors}) reveals four recurring challenges in scene-text recognition with MLLMs: (a) non-standard fonts, (b) small text size, (c) cluttered text, and (d) challenging lighting. Non-standard or stylized fonts, such as the cursive script on a restaurant sign, caused the model to misread “Dadu’s” as “DADA’s,” highlighting a lack of robustness to typography underrepresented in training data. Small text relative to image resolution often went undetected, as in the case of a car license plate “TS 08 T 4528” being read only as “4528.” In cluttered settings, dense arrangements of instructional or decorative text with varying fonts, colors, and sizes obscured the target words—for example, the fire extinguisher brand “Firewin” was recognized as “Firewion” due to surrounding label content. Lighting conditions further impacted performance: in a night-time scene, high exposure and blooming from a glowing LED sign rendered “Barbershop” illegible, with only the more prominent “Royal” partially recognized. These examples underscore the difficulty of achieving robust performance under the diverse and imperfect visual conditions.

\begin{table}[h]
\centering
\small
\caption{\textbf{Quantitative results for Task 2:} Average latency, and ANLS for each model over all currency‑note queries. The best and second-best performing models are highlighted using \textbf{bold} and \underline{underline}, respectively.}
\label{tab:currency_metrics}
\begin{tabular}{lrrr}
\toprule
\textbf{Model}         & \textbf{Latency (s)}  & \textbf{ANLS} & \textbf{LLM}\\
\midrule
BakLLaVA~\cite{liu2024improved}     & \textbf{0.92}           & 0.16         & 0.28 \\
FastVLM-7B~\cite{vasu2025fastvlm}   & 15.02                   & 0.17          & 0.37 \\
LLaVA~\cite{liu2023visual}          & 3.00                    & 0.22          & 0.32 \\
Gemma-3-27B~\cite{team2025gemma}    & 16.03                   & 0.24          & 0.98 \\
Gemma-3-12B~\cite{team2025gemma}    & 8.48                    & 0.24          & 0.58 \\
LLaVA–Phi3-3.8B~\cite{hanoona2024LLaVA++} & \underline{1.00}  & 0.24          & 0.28 \\
Qwen2.5-VL-7B~\cite{bai2025qwen2}   & 9.52                    & 0.27          & 0.44 \\
Gemma-3-4B~\cite{team2025gemma}     & 7.46              & \underline{0.27}    & 0.40 \\
LLaVA–LLaMA3~\cite{hanoona2024LLaVA++}  & 2.93           & \textbf{0.28}      & 0.30 \\

\bottomrule
\end{tabular}
\end{table}

\noindent\textbf{Task 2: Currency Recognition.} 
Each currency image was evaluated using four standardized prompts targeting distinct aspects of currency understanding: identifying the note’s value, determining its denomination and origin, describing its visual elements, and extracting any visible serial numbers or security features. As shown in Tab.~\ref{tab:currency_metrics}, LLaVA-LLaMA3~\cite{hanoona2024LLaVA++} achieved the highest ANLS (0.2787), benefiting from strong multimodal alignment. Gemma-3-4B~\cite{team2025gemma} and Qwen2.5-VL-7B~\cite{bai2025qwen2} ranked next, with Gemma-3-4B~\cite{team2025gemma} performing well despite its smaller size and Qwen’s robustness aided by fine-grained vision encoding and multilingual training. Increasing model size within the same family did not consistently improve results, underscoring that architectural alignment and task-specific visual grounding matter more than parameter count for fine-grained currency recognition.

\begin{figure}[!ht]
    \centering
    \includegraphics[width=\columnwidth]{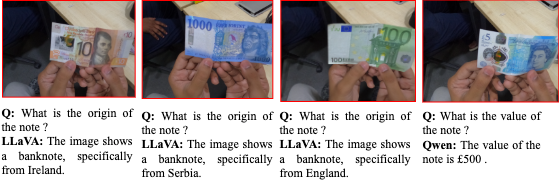}
    \caption{Qualitative examples of failure cases for currency recognition.}
    \label{fig:currency-errors}
\end{figure}

In currency recognition (see Fig.~\ref{fig:currency-errors}), MLLMs often falter at fine-grained classification—misidentifying denomination or origin even when correctly detecting a banknote. For instance, in Fig.~\ref{fig:currency-errors}, a Hungarian note was attributed to Serbia after the model failed to read “MAGYAR NEMZETI BANK” and instead may have guessed from visually similar notes in its training data. Folded or partially obscured notes can hide denomination markers, while blur from motion or poor focus reduces legibility of security details and serial numbers. These cases show the models’ dependence on clear, unobstructed text and their limited robustness to worn, folded, or poorly captured currency.

In terms of latency, BakLLaVA~\cite{liu2024improved} offers the fastest average inference at 0.92s, followed by LLaVA-Phi3-3.8B~\cite{hanoona2024LLaVA++} (1.83s) and LLaVA-LLaMA3~\cite{hanoona2024LLaVA++} (2.93s), making them appealing for real-time applications. In contrast, larger models like Gemma-3-27B~\cite{team2025gemma} (16.03s) and Qwen2.5-VL-7B~\cite{bai2025qwen2} (9.52s) exhibit significantly higher inference times, highlighting a trade-off between accuracy and responsiveness that must be considered in deployable assistive systems.

% % Exact‐match accuracy was 0\% across all models and queries, reflecting the free‑form nature of model outputs versus rigid ground‑truth strings. ANLS scores ranged from 0.1609 to 0.2787 (Table~\ref{tab:currency_metrics}), with higher values observed in models that more consistently recognized key tokens (numerals, currency names) despite descriptive phrasing. Latency varied from under 1s (BakLLaVA) to over 16s (Gemma3:27b), highlighting trade‑offs between model size, inference speed, and recognition fidelity. These results demonstrate that similarity‑based metrics are essential for evaluating open‑ended descriptive tasks and that optimized, lightweight models may offer practical advantages in real‑world assistive deployments.

% % \paragraph{Experimental Setup.}  
% % We assembled a dataset of 22 distinct banknotes spanning Indian rupees, US dollars, and Euros. Frontal photographs were captured under varied lighting conditions. Each image was queried with four standardized prompts:
% % \begin{enumerate}
% %   \item What is the value of the note?
% %   \item What is the denomination and origin of the currency?
% %   \item Describe the visual elements on the note.
% %   \item Provide any serial number or security details visible.
% % \end{enumerate}
% % We evaluated eight vision–language models (Gemma3: 4b, 12b, 27b; LLaVA; LLaVA–Llama3; LLaVA–Phi3:3.8b; Qwen2.5vl:7b; BakLLaVA) using strict exact‐match accuracy and the Average Normalized Levenshtein Similarity (ANLS) metric against canonical ground‑truth captions.

\begin{figure}[t]
    \centering
    \includegraphics[width=\columnwidth]{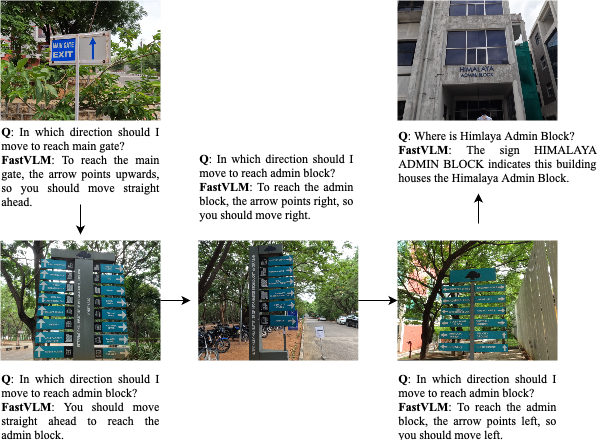}
    \caption{Example output from FastVLM~\cite{vasu2025fastvlm} for a navigation query involving the destination “Himalaya Admin Block.” The model correctly interprets scene elements like buildings and signage.}
    \label{fig:navigation_results}
\end{figure}

\noindent\textbf{Task 3: Navigation Assistance.} We also assess the directional understanding of MLLMs. To do this, we extracted a set of real-world navigation routes, as detailed in Section~\ref{sec:app}. Fig.~\ref{fig:navigation_results} presents one such qualitative example, illustrating how MLLMs interpret spatial cues and directional references within egocentric scenes. Our observations show that MLLMs can effectively guide users using navigation signs, provided the text is reliably detected and read—highlighting that overcoming the scene-text recognition challenges in Task 1 is key to unlocking their full potential in navigation assistance.

\begin{table}[h]
  \tiny
  \centering
  \caption{\textbf{Quantitative results for Task 4:} Average latency, and ANLS for each model for Understanding Indic Languages. The best and second-best performing models are highlighted using \textbf{bold} and \underline{underline}, respectively.}
  \label{tab:merged-menu-columnwise}
  \begin{tabular}{l|cc|cc|cc|c}
    \toprule
    \rowcolor{lightgray}
    \textbf{Model} & 
    \multicolumn{2}{c|}{\textbf{English}} & 
    \multicolumn{2}{c|}{\textbf{Hindi}} & 
    \multicolumn{2}{c|}{\textbf{Bengali}} &
    \textbf{Avg Latency (s)} \\
    \cmidrule{2-7}
    & \textbf{LLM} & \textbf{ANLS}  
    & \textbf{LLM} & \textbf{ANLS} 
    & \textbf{LLM} & \textbf{ANLS}  \\
    \midrule
    Gemma-3-4B~\cite{team2025gemma}          & 0.69 & 0.47 & 0.42 & 0.36 & 0.51  & \textbf{0.43} & 4.86 \\
    Gemma-3-12B~\cite{team2025gemma}         & 0.70 & 0.45 & 0.57 & \underline{0.37} & 0.67 & 0.37 & 6.11 \\
    Gemma-3-27B~\cite{team2025gemma}         & 0.96 & 0.48 & 0.83 & \textbf{0.40} & 0.93 & \underline{0.41} & 7.82 \\
    LLaVA~\cite{liu2023visual}               & 0.04 & 0.17 & 0.00 & 0.13 & 0.06 & 0.12 & 2.68 \\
    LLaVA–LLaMA3~\cite{hanoona2024LLaVA++}        & 0.22 & \underline{0.55} & 0.05 & 0.28 & 0.03 & 0.27 & 1.54 \\
    LLaVA–Phi3–3.8B~\cite{hanoona2024LLaVA++}     & 0.11 & 0.25 & 0.06 & 0.19 & 0.03 & 0.16 & 1.43 \\
    Qwen2.5–VL–7B~\cite{bai2025qwen2}       & 0.19 & \textbf{0.59} & - & - & -  & -   & - \\
    BakLLaVA~\cite{liu2024improved}            & 0.11 & 0.11 & 0.03 & 0.12 & 0.01 & 0.06 & 0.93 \\
    FastVLM-7B~\cite{vasu2025fastvlm}             & 0.12   &   0.18    & 0.12     &     0.16   &   0.04    &     0.09 & 13.42 \\
    \bottomrule
  \end{tabular}
\end{table}

% \subsection{Evaluation of Understanding Indic Languages}

\begin{figure}[!h]
    \centering
    \includegraphics[width=\columnwidth]{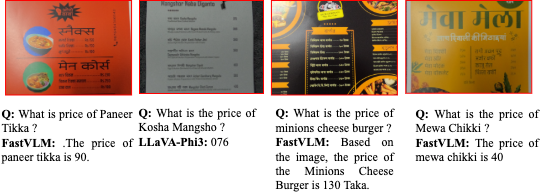}
    \caption{Qualitative examples of failure cases for multilingual menu card reading and interpretation.}
    \label{fig:menu-errors}
\end{figure}
\noindent\textbf{Task 4:  Multilingual Menu Card Reading and Interpretation.} Table~\ref{tab:merged-menu-columnwise} summarizes the performance of various MLLMs on understanding restaurant menu cards across English, Hindi, and Bengali. Notably, among all evaluated models, only the Gemma family was explicitly trained with support for Indic languages; the others were primarily trained on English and high-resource languages. Despite this, Qwen2.5-VL-7B~\cite{bai2025qwen2} achieves the highest ANLS score on English menus (0.5588), followed closely by LLaVA-LLaMA3 (0.5460) and Gemma-3-27B (0.4807). For Hindi, Gemma-3-27B~\cite{team2025gemma} outperforms all other models with an ANLS of 0.4035, with Gemma-3-12B~\cite{team2025gemma} (0.3705) as the second-best. Surprisingly, for Bengali, the smaller Gemma-3-4B~\cite{team2025gemma} performs best (0.4249), outperforming its larger siblings. This result suggests that multilingual pretraining in Gemma~\cite{team2025gemma} helps significantly, especially for scripts and content not represented in the training distributions of other models. Due to the generation of non-decodable or illegal Unicode characters by Qwen2.5-VL-7B~\cite{bai2025qwen2} in Indic language outputs, we were unable to include its results in the Hindi and Bengali evaluations.

In the menu card example (refer Fig.~\ref{fig:menu-errors}), the model struggles to accurately identify and associate dish names with their corresponding prices due to a combination of challenges: visually complex layouts with multiple columns, varying background colors, suboptimal lighting or exposure, and the simultaneous presence of multiple languages. Such conditions can confuse text localization, hinder script recognition, and break the logical association between related text elements. These limitations significantly reduce the robustness of MLLMs for real-world menu card understanding, especially in visually cluttered and multilingual environments. Gemma-3-27B~\cite{team2025gemma} attains the highest LLM-as-a-judge performance~\cite{li2024llms}, with accuracies of 96.29\%, 83.11\%, and 93.13\% for English, Hindi, and Bengali, respectively.

In terms of latency, lightweight models such as BakLLaVA~\cite{liu2024improved} and LLaVA-Phi3-3.8B~\cite{hanoona2024LLaVA++} demonstrate the fastest inference across all languages, with BakLLaVA~\cite{liu2024improved} consistently achieving sub-1s latency, making it ideal for low-latency applications. However, this speed comes with a trade-off BakLLaVA~\cite{liu2024improved} shows very low ANLS scores, reflecting weak text recognition capabilities. While Qwen2.5-VL-7B~\cite{bai2025qwen2} achieves high accuracy on English, it lacks multilingual capacity and was not evaluated on Hindi or Bengali. Interestingly, FastVLM-7B~\cite{vasu2025fastvlm}, despite its branding, exhibits significantly higher latency (12–14s) while delivering only moderate accuracy, suggesting that its architecture is optimized for different tasks. Overall, the results highlight that Indic-language performance is highly dependent on pretraining exposure, and that models not trained on regional scripts suffer both in accuracy and consistency—underscoring the value of multilingual pretraining for real-world assistive AI deployments in linguistically diverse regions.

\noindent\textbf{Task 5: Reading Long-form Content} 
We evaluated four MLLMs — DeepSeek-R1-14B~\cite{guo2025deepseek}, LLaMA3.1-8B~\cite{dubey2024llama}, Phi-4-14B~\cite{abdin2024phi}, and Qwen3-14B~\cite{yang2025qwen3} — measuring inference latency and ANLS (Table~\ref{tab:book-reading-results}). LLaMA3.1-8B~\cite{dubey2024llama}~\cite{guo2025deepseek} achieved the best balance with the highest ANLS (0.1278) and low latency (13.34s). DeepSeek-R1-14B~\cite{guo2025deepseek} followed in accuracy but was slower (56.64s). Qwen3-14B~\cite{yang2025qwen3} and Phi-4-14B~\cite{abdin2024phi} underperformed in accuracy and speed, likely due to limitations in long-context reasoning. These results highlight the need for models that are both context-aware and latency-efficient for assistive reading tasks: average  latency and ANLS for each model on the Book Reading QA task. Latency reflects the time to process a single query, showing large differences across models—from 13.34s for LLaMA3.1-8B~\cite{dubey2024llama} to over 130s for Qwen3-14B~\cite{yang2025qwen3}—highlighting the trade-off between speed and accuracy in real-time reading assistance. The LLM scores are relatively low overall (0.22–0.29), underscoring the difficulty of the task. This suggests that, despite similar ANLS values, models differ in how often their responses are judged semantically adequate by another LLM.

\begin{table}[h]
  \centering
  \small
  \caption{\textbf{Quantitative results for Task 5:} Average latency, exact‑match accuracy, and ANLS for each model on the Book Reading QA task.}
  \label{tab:book-reading-results}
  \begin{tabular}{lrrr}
    \toprule
    \textbf{Model}         & \textbf{Latency (s)}  & \textbf{ANLS} & \textbf{LLM}\\
    \midrule
    DeepSeek-R1-14B~\cite{guo2025deepseek}   & 56.64           & 0.12         & 0.28 \\
    LLaMA3.1-8B~\cite{dubey2024llama}         & 13.34          & 0.13         & 0.24 \\
    Phi-4-14B~\cite{abdin2024phi}                & 30.57       & 0.06         & 0.22 \\
    Qwen3-14B~\cite{yang2025qwen3}           & 130.07          & 0.07         & 0.29 \\
    \bottomrule
  \end{tabular}
\end{table}

Our findings reveal two key challenges: in multilingual scenarios (Task 4), limited script exposure leads to basic recognition failures (e.g., LLaVA, Qwen) or misalignment in complex layouts (e.g., Gemma). In long-context reading (Task 5), accumulated OCR errors and weak reasoning across sequential inputs hinder models that handle single-image VQA well but fail to maintain narrative consistency across pages. We present further qualitative examples for all tasks in the supplementary.

\section{Conclusion}

In this work, we explored the potential of MLLMs for a spectrum of visual assistive tasks, including scene-text recognition, currency note identification, navigation assistance, multilingual menu reading, and long-form content access. Our results show that, when coupled with speech and question-answering interfaces, MLLMs offer a promising route toward AI assistants that substantially improve accessibility for people with visual impairments. Such systems can deliver real-time, context-aware information, helping bridge the gap between the visual world and non-visual perception. Our findings generalize broadly, as the tasks test core abilities—text recognition, navigation, and object understanding—while exposing real-world challenges like poor lighting, small fonts, complex layouts, multilingual text, and contextual reasoning.

At the same time, our experiments reveal clear limitations: performance degrades in complex, real-world scenarios with multi-column layouts, low light, cluttered backgrounds, and multilingual content, underscoring gaps in robustness. Future work should therefore focus on domain- and task-specific fine-tuning, as well as on designing more natural, accessible interfaces tailored to visually challenged users. Reducing latency will also be crucial, since slow responses disrupt interaction and undermine usability in time-sensitive situations. With these targeted improvements, MLLMs can progress from promising prototypes to dependable assistive technologies suitable for everyday use.

\section{Acknowledgments}
Shayon Dasgupta is supported through the SRFP by the Indian Academy of Sciences. Avijit Dasgupta is supported by a Google Ph.D. India Fellowship. The work is supported by MeitY, Govt. of India, through the NLTM-Bhashini project.

\bibliographystyle{ACM-Reference-Format}
\bibliography{ICVGIP-Latex-Template}

% \appendix

\end{document}